\documentclass[conference]{IEEEtran}
\IEEEoverridecommandlockouts

\usepackage[numbers,sort&compress]{natbib}
\usepackage{amsmath,amssymb,amsfonts}
\usepackage{float} 
\usepackage{subfigure} 
\usepackage{algorithm} 
\usepackage{algpseudocode}
\usepackage{times}
\usepackage{epsfig}
\usepackage{graphicx}
\usepackage{xcolor}
\usepackage{listings}
\usepackage{nicefrac,xfrac}
\usepackage{etoolbox}
\usepackage{tabulary}
\usepackage[british,american]{babel}
\usepackage{diagbox}
\usepackage{enumitem}
\usepackage{balance}
\newcommand{\dist}{\mathcal{D}}
\newcommand{\p}{{p}}  
\newcommand{\q}{{q}}  

\newcommand{\lnorm}[1]{\frac{#1}{\left\lVert{#1}\right\rVert _2}}

\usepackage{xspace}
\makeatletter
\DeclareRobustCommand\onedot{\futurelet\@let@token\@onedot}
\def\@onedot{\ifx\@let@token.\else.\null\fi\xspace}

\def\etc{\emph{etc}\onedot} \def\vs{\emph{vs}\onedot}

\makeatother

\title{Augmentation-induced Consistency Regularization for Classification}

\DeclareRobustCommand*{\IEEEauthorrefmark}[1]{
	\raisebox{0pt}[0pt][0pt]{\textsuperscript{\footnotesize\ensuremath{#1}}}}
\begin{document}

\author{\IEEEauthorblockN{Jianhan Wu\IEEEauthorrefmark{1,2}, Shijing Si\IEEEauthorrefmark{1*}, Jianzong Wang\IEEEauthorrefmark{1} , Jing Xiao\IEEEauthorrefmark{1}}
	\IEEEauthorblockA{\emph{\IEEEauthorrefmark{1}Ping An Technology (Shenzhen) Co., Ltd.,} Shenzhen, China\\
		\IEEEauthorblockA{\emph{\IEEEauthorrefmark{2}University of Science and Technology of China,} Hefei, China  }
		Emails: wujianhan@mail.ustc.edu.cn, shijing.si@outlook.com, jzwang@188.com, xiaojing661@pingan.com.cn}
	\thanks{* Corresponding author: Shijing Si, shijing.si@outlook.com}
}
\maketitle

\begin{abstract}
Deep neural networks have become popular in many supervised learning tasks, but they may suffer from overfitting when the training dataset is limited. To mitigate this, many researchers use data augmentation, which is a widely used and effective method for increasing the variety of datasets. However, the randomness introduced by data augmentation causes inevitable inconsistency between training and inference, which leads to poor improvement. In this paper, we propose a consistency regularization framework based on data augmentation, called CR-Aug, which forces the output distributions of different sub models generated by data augmentation to be consistent with each other. Specifically, CR-Aug evaluates the discrepancy between the output distributions of two augmented versions of each sample, and it utilizes a stop-gradient operation to minimize the consistency loss. We implement CR-Aug to image and audio classification tasks and conduct extensive experiments to verify its effectiveness in improving the generalization ability of classifiers. Our CR-Aug framework is ready-to-use, it can be easily adapted to many state-of-the-art network architectures. Our empirical results show that CR-Aug outperforms baseline methods by a significant margin. 
\end{abstract}
\section{Introduction}
Deep Learning (DL) methods have received wide attention due to their powerful capability and good performance. Many deep neural network (DNN) architectures, including  ResNet \cite{he2016deep}, Transformer \cite{vaswani2017attention}, BERT \cite{DBLP:conf/naacl/DevlinCLT19} and gMLP \cite{liu2021pay}, have shown great promise in computer vision (CV), audio processing and natural language processing (NLP). However, most models rely on a large amount of labeled data to learn parameters when training or fine-tuning \cite{shorten2019survey}, which may subject to overfitting or weak generalization performance in low-resource datasets.

To alleviate overfitting and enhance the generalization ability of deep learning models, many methods have been proposed, such as data augmentation \cite{gong2021keepaugment}, dropout \cite{srivastava2014dropout}, batch normalization \cite{ioffe2015batch}, weight decay \cite{Loshchilov2019DecoupledWD}, pretraining \cite{erhan2010does}, variational information bottleneck \cite{Si2021VariationalIB}, etc. The most recently developed method is R-Drop \cite{liang2021r}, which forces the output distributions of two different sub-networks generated from dropout by utilizing Kullback-Leibler (KL) divergence \cite{yu2013kl} in the training stage to be consistent. And R-drop achieves substantial improvements on 5 NLP and CV tasks by adding a KL-divergence loss without modifying the structure.

Data augmentation is a very common method to mitigate model overfitting for its effectiveness, and has good compatibility with other regularization methods. To the best of our knowledge, most researchers only focus on studying how to design good data augmentation methods for certain tasks at the data level, and almost no one studies data augmentation at the model output level. Inspired by R-Drop, we argue that a good DNN-based classifier should yield predictions that are invariant to multiple data augmentations since data augmentation simply morphs the data without affecting its character. Furthermore, there are a variety of augmentation strategies for different data that will create useful sampled data for the regularized DNN model rather than generating random model parameters like R-Drop. As a result, we propose that data augmentation-based regularization can constrain deep models more effectively than R-Drop, particularly in datasets with limited resources.

In this paper, we develop a framework, called CR-Aug, short for consistency regularization based on data augmentations. Specifically, our framework proceeds as follows. Firstly, for each input sample, we perform data augmentation twice to obtain a pair of augmented samples; then we feed the pair to the deep encoder, which predicts two output distributions. 
Subsequently, we compute the consistency loss between the two output distributions for each pair of augmented data, which forms the total loss with cross-entropy loss of the pair. When optimizing the loss, we implement stop-gradient operation to the consistency loss to avoid collapsing solutions. Stop-gradient operation has been shown beneficial to the training of DNNs in many aspects, for instance, in Siamese representation learning \cite{chen2021exploring} and in a multi-task setting \cite{zhang2021symmetric}. Experiments show that CR-Aug outperforms several regularization methods (such as R-Drop), demonstrating that it is a good regularization method capable of alleviating the overfitting problem in classification.

Though our CR-Aug framework is straightforward, we find it surprisingly effective through experiments on both image and audio datasets.
 
Our main contributions can be summarized as follows:
\begin{itemize}[itemsep=8pt,topsep=8pt,parsep=0pt,partopsep=0pt,leftmargin=15pt]
	\item We propose CR-Aug, a simple yet effective data augmentation regularization framework, which can be widely used in deep learning classification tasks.
	\item We conduct extensive experiments on both audio and image datasets with two popular network architectures to verify the effectiveness of our method in comparison with other baseline methods.
	\item A comprehensive ablation study reveals that stop-gradient operation and consistency loss function plays a critical role in the performance of CR-Aug.
\end{itemize}
\section{related works}
\subsection{Data Augmentation Regularization Methods}
Deep models bring better performance as moving toward large parameter models, especially some large pre-trained models like BERT, Vision Transformer \cite{dosovitskiy2020image}, GPT family \cite{radford2018improving, floridi2020gpt}, \etc But with that comes overfitting when the training dataset is limited, which requires regularization to reduce it. At present, there are some regularization methods, such as batch normalization, weight decay, \etc Among them, the most commonly used method is data augmentation, which is widely used successfully in many neural network architectures, such as convolutional neural networks (CNN), recurrent neural networks (RNN), Transformer. Its success can be interpreted as randomly cropping, scaling, and rotating the data in the data dimension to increase the training data and make the training data as diverse as possible, which makes the trained model have stronger generalization ability. For images, a recently developed method is random erasing \cite{zhong2020random}, which randomly selects a rectangular area and erases pixel values with random numbers. Doing so can get a lot of occluded training data, which is helpful for training. Therefore, the trained model is able to mitigate overfitting and be robust to occluded images. For audio, a recently developed method is the ensemble of different data augmentation \cite{nanni2020data}, which improves the diversity of training data by integrating multiple audio data augmentation methods. Compared with the method without data augmentation, the accuracy of this method on animal audio classification is greatly improved. 

Unlike previous studies that sought to design specifically data augmentation methods to improve classification accuracy, we further investigate regularization models in the context of the success of data augmentation. Specifically, by using cosine similarity to drive data-augmented sub-models for each input data to produce similar predictions, that is, we perform a regularization operation at the model output level. Doing so not only increases the randomness of the data but also reduces the parameter freedom of the sub-model, which improves the generalization performance of the model during the inference phase.
\subsection{Consistency Regularization}
	The main idea behind consistency regularization is that its predictions should be consistent with small perturbations in the input \cite{abuduweili2021adaptive}. This method is widely used in semi-supervised and self-supervised learning because it can learn a good representation of the input data \cite{sinha2021consistency}. The perturbation is generally performed on the model and the data, and dropout is commonly used for the model. FD \cite{zolna2017fraternal} trains two copies of the same model with different dropout masks, using $L_2$ loss to minimize the difference between their predictions, this method performs well in both NLP and CV. Data augmentation is frequently used at the data level. Cutoff \cite{shen2020simple} accomplishes the goal of augmenting data by deleting some pixels of the data. Then performing Consistency training on both the original and augmented data. UDA \cite{xie2020unsupervised} emphasizes regularization training on semi-supervised learning using the state-of-the-art data augmentation methods. However, these methods only focused on the data augmentation method and did not discover the link between data augmentation and consistency regularization. We propose CR-Aug, which employs cosine similarity to constrain data that has undergone multiple data augmentation and sampling to generate consistency predictions to mine the relationship of data augmentation with consistency regularization. The effectiveness of CR-Aug as a consistency regularization method is demonstrated by theoretical analysis and experimental verification.
\section{Methodology}

In this section, we elaborate on the details of how CR-Aug works. Firstly, we detailedly introduce the architecture and the algorithm of our framework and explain its principles. Then we cover the objective loss functions in our paper.

\begin{figure}[h] 

\centering 
\includegraphics[height=0.175\textwidth,width=0.49\textwidth]{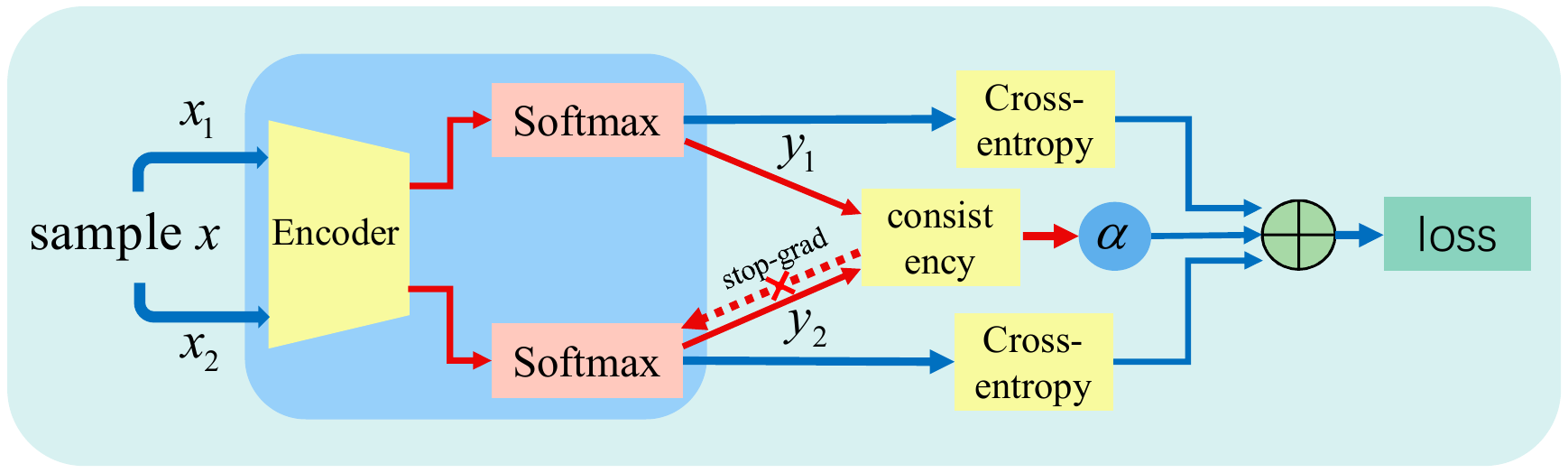}
\caption{The overall architecture of our CR-Aug framework for classification.} 
\end{figure}
\subsection{CR-Aug}\label{aug}
The overall architecture of CR-Aug is shown in Figure \ref{Fig.main1}. For each input data $x$, we first perform data augmentation to obtain $x_1$ and $x_2$, and then feed these two augmented data into the same deep neural networks (DNN) backbone (an encoder, such as ResNet18). Subsequently, we have the logits from the encoder and pass them to a Softmax layer to yield the output distributions ${y}_1$ and ${y}_2$. Finally, we calculate the consistency loss between the distributions ${y}_1$ and ${y}_2$, plus the cross-entropy loss between the output distribution and ground-truth label $y$. 

Following \cite{chen2021exploring}, we utilize the stop-gradient operation to prevent collapsing when minimizing the consistency loss. For clarity, the pseudo-code of CR-Aug is presented in Algorithm \ref{alg:code}. The key of CR-Aug lies in the consistency loss between the output distributions of two augmented samples for each input sample. Essentially, this loss ensures that a good encoder should produce output distribution invariant to the data augmentations.

Introducing the consistent loss may yield insignificant improvement in model performance. Therefore, we deploy stop-gradient operation (the detach function in the Algorithm \ref{alg:code}), which is shown useful to the training of DNNs in the literature. 

The specific principle of how stop-gradient works is as follows: the purpose of our consistency $loss_2$ is to constrain the network to learn meaningful representations, which can be defined as the following form:
\begin{equation}
	L(\theta ,\eta )=\mathbb{E}_{x,\Gamma }[\parallel {F_\theta }(\Gamma (x)) - {\eta _x}{\parallel _2}]
\end{equation}
Where $\Gamma $ represents data augmentation, ${{\rm{F}}_\theta }$ represents the encoder function, and ${\eta _x}$ represents the representation of the input x. $\mathbb{E}$ is the expectation of x and data augmentation (that is, the sum of the expected loss of all inputs and data augmentation). For the convenience of analysis, the equivalent form of cosine similarity MSE \cite{kang2020constrained} is used to represent the similarity. Then the optimization objective can be simplified as:
\begin{equation}
	{\min _{\theta ,\eta }}L(\theta ,\eta )
\end{equation}
\begin{algorithm}[t]

\caption{CR-Aug, PyTorch-like}
\label{alg:code}
\definecolor{codeblue}{rgb}{0.2,0.8,0.1}
\definecolor{codekw}{rgb}{0.84, 0.17, 0.49}
\lstset{
	backgroundcolor=\color{white},
	basicstyle=\fontsize{8.5pt}{9.5pt}\ttfamily\selectfont,
	columns=fullflexible,
	breaklines=true,
	captionpos=b,
	commentstyle=\fontsize{7.5pt}{7.5pt}\color{codeblue},
	keywordstyle=\fontsize{7.5pt}{7.5pt}\color{codekw},
}
\begin{lstlisting}[language=python]
  # f: backbone(ResNet50, gMLP and other models)
  # Loss1: CrossEntropyLoss
  # Loss2: Consistency loss

  for x, y in loader:  # load a minibatch x with n samples
  x1, x2 = aug(x), aug(x)  # random augmentation
  y1, y2 = f(x1), f(x2)  # n-by-d

  L = Loss1(y1, y) + Loss1(y2, y) + alpha*Loss2(y1, y2.detach())   #loss  and stop-gradient
  L.backward()  # back-propagate
  update(f)  # SGD update
\end{lstlisting}
\end{algorithm}
The form of this optimization objective is similar to the k-means \cite{krishna1999genetic} algorithm. Among them, the variable $\theta $ is similar to the cluster center, which are the learnable parameters of the encoder. The variable $\eta $ is similar to the embedding vector of the sampling point, which is the representation of the input x. Then CR-Aug can be solved by an EM iterative algorithm \cite{mclachlan2007algorithm} like the k-means algorithm, fixing one variable and estimating another variable. Formally, it can be written as the following two sub-problems:
\begin{equation}
	{\theta ^t} \leftarrow \arg \mathop {\min }\limits_\theta  L(\theta ,{\eta ^{t - 1}})\label{eq:theta}
\end{equation}
\begin{equation}
	{\eta ^t} \leftarrow \arg \mathop {\min }\limits_\eta  L({\theta ^t},\eta )\label{eq:eta}
\end{equation}

Where t represents the iteration round. It can be solved by SGD \cite{bottou2012stochastic} or ADAM \cite{kingma2014adam}. In this process, stop-gradient is employed, that is, using detach function on $y_2$. The detach function will cut the data and no backward propagation would be conducted, which plays the role of fixing ${\eta ^{t - 1}}$. If there is no stop-gradient, equation \eqref{eq:theta} has two variables and cannot be solved, which will result in collapsing solutions and bring poor results. 

After solving $\theta$, there is only one variable left in equation \eqref{eq:eta}. Substituting $\theta$ into the loss function, the second sub-problem becomes:
\begin{equation}
	{\eta ^t} \leftarrow \arg \mathop {\min }\limits_\eta \mathbb{E}_\Gamma [\parallel {F_{{\theta ^t}}}(\Gamma (x)) - {\eta _x}{\parallel _2}]
\end{equation}
Finally, the image x representation that is vital for downstream tasks after t iterations can be obtained by solving:
\begin{equation}
	\eta _x^t \leftarrow \mathbb{E}_\Gamma [{F_{{\theta ^t}}}(\Gamma (x))]
\end{equation}

It is worth noting that the effect of stop-gradient operation on any output is essentially unchanged (y2 in our experiment) since the encoder part of the model is the same (shared parameters). Besides, our experiments also verify the effectiveness of stop-gradient in the CR-Aug framework later.
\subsection{Loss Function}
Here we cover the objective functions in the CR-Aug framework.
The consistency loss between output distributions measures the discrepancy between two categorical distributions. Usually, the following three kinds of loss are commonly used: cosine similarity, KL-divergence and Jenson’s Shannon (JS) divergence. 

The cosine similarity in our experiment is defined as follows:
\begin{equation}
	\dist(p, q) = 1- \lnorm{\p}{\cdot}\lnorm{\q},
	\label{eq:dist_cosine}
\end{equation}
where ${\left\lVert{\cdot}\right\rVert _2}$ is $\ell_2$-norm. This is equivalent to the mean squared error of $\ell_2$-normalized vectors \cite{grill2020bootstrap}. And the value of the cosine similarity is between 0 and 2 after adding 1. 

Kullback-Leibler (KL) divergence and Jenson Shannon (JS) divergence \cite{fuglede2004jensen} are defined as follows:
\begin{equation}
	KL(p||q)=\int p(x)\log(\frac{p(x)}{q(x)})dx,
\end{equation}
\begin{equation}
	JS(p||q)=\frac{1}{2}KL(p||\frac{p+q}{2})+\frac{1}{2}KL(q||\frac{p+q}{2})
\end{equation}	
where $p$ and $q$ are two output distributions. It should be noted that JS divergence is a variant of KL divergence, which solves the problem of asymmetry of KL divergence.

We calculate the cross-entropy loss of the two output distributions (${y}_1$, ${y}_2$) with the real label ($y$) for classification, and calculate the consistency loss of $y_1 $ and $ y_2$ for consistency training. Therefore for each input sample, the total loss function is
\begin{equation}
	L = L_{CE}(y_1, y) + L_{CE}(y_2, y) + \alpha\cdot{L}_{Con}(y_1, y_2),
\end{equation}
where the ${L}_{Con}(y_1, y_2)$ is the consistency loss between $y_1$ and $y_2$, and it can be chosen from Cosine distance, KL divergence and JS distance. The $\alpha$ is the weight of consistency loss.

	\section{Experiment Sutup}
	\vspace{0.3cm}
	In this section, we describe how we conduct experiments to verify the superiority of our CR-Aug framework in comparison with other baseline methods. Also, we present an extensive ablation study to show the source of the improvements. 
	
	\subsection{Configuration}
	We used the Pytorch framework to conduct experiments on two classification tasks, including one image dataset and two audio datasets. For image classification, we used the ResNet18 and gMLP model as the backbone of our structure. For audio classification, we used ResNet50 as the backbone of our structure. For comparison, we set a few baseline methods are as follows:

	\begin{itemize}[itemsep=5pt,topsep=6pt,parsep=0pt,partopsep=0pt,leftmargin=15pt]
		\item\textbf{Without data augmentation (w/o Aug)}: This is a generic classification with just one cross-entropy loss. And there is no augmentation and regularization method used in this method.
		\item\textbf{R-Dropout}: R-Dropout is a recently developed method, which forces the output distributions of different sub-models generated by dropout to be consistent with each other. It can be seen as an enhanced variant of dropout. And we use the rates of dropout with the same value (0.3, 0.3). See \cite{liang2021r} for details.
		\item\textbf{Weight Decay}: This is a common regularization technique by adding a small penalty to improve generalization \cite{krogh1992simple}. We implement the weight decay of 0.1 in the SGD optimizer. 
	\end{itemize}
	In order to better compare the effect of various regularization methods, we remove all the tricks and regularization of the original state-of-the-art method. This is the reason why the accuracy of the latter baseline method is lower than that of the existing methods.

	\textbf{Metrics} For both the image and audio datasets, because the datasets are balanced, we utilize accuracy as the performance metric.
	
	\textbf{Training details} In the experiment phase, for both CR-Aug and baseline methods, the data is divide into training set, validation set and test set at a ratio of 8:1:1. The SGD  algorithm is employed as the optimizer with the default parameters. The model is trained with a learning rate of 0.001 for CIFAR-10 and 0.01 for other datasets. Batch size is set to 32 and training epoch is set to 60 for all datasets. And we use Tesla-V100 with 16G memory to experiment.
	
	\subsection{Data Augmentation}
	Because we use audio and image datasets in our experiments, we introduce data augmentation methods on audio and images separately.
	
	\subsubsection{data augmentation for images}
	For images, there are many kinds of data augmentation methods and they can be directly accessed in the Pytorch framework. Here are the augmentations we used: 1.) Flip \cite{loening2015increased}: randomly flipping the image 5 degrees in our experiments. Do not set more than 10 degrees, it will lead to poor results otherwise; 2.) Adding Gaussian noise (AN) \cite{shorten2019survey}: superimposing random noise with Gaussian distribution on pixels; 3.) color jitter \cite{takahashi2018ricap}: randomly change the exposure, saturation and hue of the image to form pictures under different lighting and colors; 4.) Randomly Resizing Cropping (RRC) \cite{shorten2019survey}: specifically, first expanding the size of the image data to $224\times224$, and then randomly cropping it. Note that the cropping ratio here cannot be too large, otherwise it will lead to lower model accuracy. In our experiment, we used a size of 0.2*1.0. For obtaining more abundant training data, we combined two or more methods above introduced on each mini-batch to obtain different augmentation schemes, such as the combination of RRC and Flip.
	
	\subsubsection{data augmentation for audio}
	There are many commonly used data augmentation methods for audio datasets, such as, time stretch (TS) \cite{ko2015audio}, pitch shifting (PS) \cite{ko2015audio}, adding Gaussian noise (AN) \cite{ko2015audio}, random masking (RM) \cite{ko2017study}, add impulse response \cite{bryan2020impulse} and so on. In our experiments, we chose four commonly used data augmentation methods: TS, PS, AN, and RM. The method TS is the slowdown or speed up the audio sample while keeping the pitch unchanged. The method PS is that raise or lower the pitch of the audio sample without changing the duration. The method AN is that randomly adds Gaussian noise to the audio sample. The method RM is to randomly set the value in a few percent (five in our experiments) of audio samples to 0. In order to generate more diverse data augmentation schemes and obtain more abundant training data, we also combined two or more of the above methods, which could get more satisfactory results in the next experiments.
	\section{Experimental Results}	
	\vspace{0.3cm}
	\subsection{Application To Image Classification}
	\vspace{0.1cm}
	\textbf{Dataset and Backbone} For image classification, we conduct experiments on one widely used benchmark dataset, that is CIFAR-10 \cite{krizhevsky2009learning} dataset. CIFAR-10 dataset consists of 60000 $32\times32$ images of 10 classes, and there are 6000 images per class. We divide these 60000 images into training set, validation set and test set at a ratio of 8:1:1. We used two different DNN models (ResNet18 and gMLP \cite{liu2021pay}) as the backbone of our framework in the experiments.  
	
	\textbf{Results and Analysis} Table \ref{table1} displays the classification accuracy of various methods on CIFAR-10 dataset.
	The method with mixed augmentation (we marked it as Mixed Aug) outperforms the w/o aug baseline method by about 16\% on gMLP backbone. The Mixed Aug method is the combination of multiple data augmentation methods (including RRC, flip, and color jitter), which brings rich images to make CR-Aug robustly help improve the model generalization and model performance. And it can be seen that RRC is the most effective single data augmentation method. R-dropout does not perform better than weight decay in ResNet backbone for that dropout is not suitable for applying to ResNet \cite{he2016identity}. Noting that our results achieve 93.41\% accuracy for ResNet-18 on CIFAR-10, which is lower than the state-of-the-art results. This is because our model uses only data augmentation without any other regularization methods, whereas the state-of-the-art model uses lots of tricks and regularization methods that can improve generalization performance.
	By comparing the baseline method and our methods (RRC, flip, AN, RRC+flip, and Mixed Aug), we not only concluded the superiority of our method but also discovered the combination of several data augmentation methods work better than single data augmentation for image classification.
	\begin{table}[h]\centering
		\renewcommand{\arraystretch}{1.4}
		\setlength{\tabcolsep}{5mm}
		\caption{The classification accuracy of various methods on CIFAR-10 dataset using two network backbones. RRC: random resize croping. RD: R-dropout. AN: add Gaussian noise. Mixed Aug: the best combination of data augmentations.}\label{table1}
		\begin{tabular}{l| c | c} 
			\hline
			\diagbox[width=3.4cm, height=0.9cm]{Method}{Backbone} & ResNet18 &  gMLP \\
			\hline
			w/o aug & 81.71$\pm$0.07 & 68.42$\pm$0.32 \\
			+RD & 82.13$\pm$0.16&  70.65$\pm$0.22 \\
			+Weight Decay  & 83.96$\pm$0.31 & 70.64$\pm$0.05 \\
			\hline
			+RRC & 91.71$\pm$0.15 & 79.18$\pm$1.25\\
			
			+flip & 84.88$\pm$0.56& 74.72$\pm$0.08 \\
			+AN & 85.95$\pm$0.12& 71.58$\pm$0.58  \\
			+RRC+flip & 93.24$\pm$0.11 & 80.06$\pm$0.75  \\
			\textbf{+Mixed Aug} & \textbf{93.41$\pm$0.27} & \textbf{84.14$\pm$0.43} \\
			\hline
		\end{tabular}
	\end{table} 

	\subsection{Application To Audio Classification}
	\vspace{0.1cm}
	\textbf{Dataset and Backbone} We conduct experiments on two audio datasets to verify the wide applicability of our framework. That is the Google Speech Commands dataset v0.01 \cite{warden2017speech} and the Audio-MNIST dataset \cite{kim2020digital}. The Google Speech Commands dataset has 64720 one-second long utterances of 30 short words. These words are from a small set of commands and are spoken by a variety of different speakers. We load the data at a sampling rate of 16000. The Audio-MNIST consists of 30000 audio samples of spoken digits (0-9) of 60 different speakers \cite{kim2020digital}. Every sample is only 0.5-seconds-long. We use it to do 60 classification tasks. And ResNet50 as the backbone of our framework.
	
	\textbf{Results and Analysis} Table \ref{table2} displays the classification accuracy of various datasets on Resnet50 backbone. The results show that our methods (TS, PS, AN, RM, TS+PS, and Mixed Aug) can improve the model's accuracy on the Speech Commands dataset and Audio-MNIST dataset, exhibiting improved generalization performance. Since different data augmentation methods have different effects on different datasets, we only find the +RM method can outperform all baseline methods (but all higher w/o aug). 
	But the accuracy of the combination of multiple data augmentation methods is better than baseline methods. And the best result Mixed Aug method (including RM, TS, and AN) achieved more than 1.2\% accuracy than R-drop baseline on the Speech Commands dataset, which proves that our CR-Aug is a practical and ready-to-use regularization method.

	\section{Ablation Study}
	\vspace{0.2cm}
	To explore the source of the effect of CR-Aug on mitigating overfitting, We did an extensive ablation study to analyze the effect of each part of our framework on improving accuracy. For image classification, we use ResNet18 as our backbone, and the dataset is CIFAR-10, the data augmentation method is flip. For audio classification, we use ResNet50 backbone and Audio-MNIST dataset to conduct experiments. And the data augmentation method is RM. The influence of stop-gradient, loss function, and consistency loss coefficient $\alpha$ on accuracy are analyzed in detail as follows. 
	\begin{table}[!h]\centering
		\vspace{-0.2cm}
		\renewcommand{\arraystretch}{1.4}
		\setlength{\tabcolsep}{1mm}
		\caption{The classification accuracy on two audio datasets with ResNet50 backbone. TS: time stretching. PS: pitch shifting. AN: add Gaussian noise. RM: random mask. RD: R-dropout. Mixed Aug: the best combination of data augmentation}\label{table2}
		\begin{tabular}{l| c| c}
			\hline
			\diagbox[width=3.6cm, height=1.0cm]{Method}{Dataset} & Speech Commands & Audio-MNIST\\
			\hline
			w/o aug & 94.12$\pm$1.52 & 95.45$\pm$0.27  \\
			+RD   & 96.52$\pm$0.99 & 98.35$\pm$0.32 \\
			+Weight Decay  &  96.65$\pm$0.16 & 99.36$\pm$0.21 \\
			\hline
			+TS & 96.07$\pm$0.55 &  98.96$\pm$0.13 \\
			+PS & 96.25$\pm$0.66 & 99.06$\pm$0.21 \\
			+AN & 96.02$\pm$0.82 & 98.62$\pm$0.33 \\
			+RM & 96.66$\pm$0.58 & 99.41$\pm$0.21 \\
			+TS+PS & 96.63$\pm$0.81 & 99.35$\pm$0.18 \\
			+Mixed Aug & \textbf{97.21$\pm$0.69} & \textbf{99.51$\pm$0.18}  \\
			\hline
		\end{tabular}
			\vspace{-0.3cm}
	\end{table}
	\subsubsection{the effect of the stop-gradient} \label{stop-grad}
\begin{figure}[h] 
		\centering 
		\includegraphics[width=0.53\textwidth]{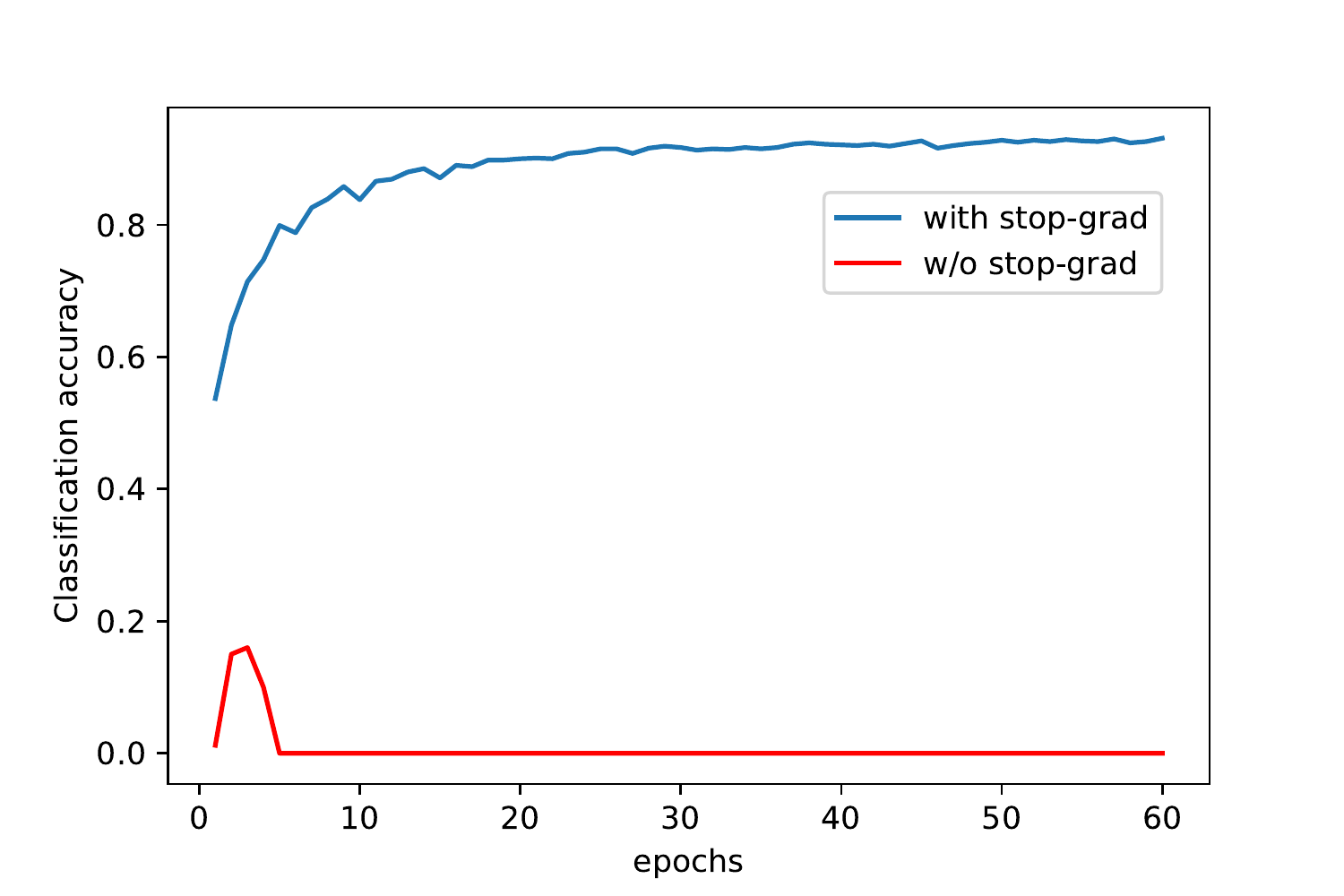} 
		\caption{Collapsing solutions \vs Normal.} 
		\label{loss} 
	\end{figure}

	During our experiment, we found that the classification accuracy of the model suddenly becomes very low (almost zero). We argue that the gradient disappeared or collapsing solutions appeared. Inspired by \cite{chen2021exploring}, we used the stop-gradient to deal above situations. We did an ablation study to prove the effect of introducing stop-gradient. As shown in figure \ref{loss}, The classification accuracy becomes 0 after 5 epochs, which is not the case for experiments with stop-gradient. As quantified in Table \ref{table3}, even if there is no collapse solution, stop-gradient can improve the classification accuracy. The results show that the introduction of stop-gradient can not only mitigate the collapsing solutions but also improve the accuracy of classification tasks. And the principle can be seen in the section \ref{aug}.
	\vspace{-0.2cm}
	\begin{table}[H]\centering
		\renewcommand{\arraystretch}{1.4}
		\caption{The results of effect of stop-gradient.}\label{table3}
		\setlength{\tabcolsep}{3mm}
		\begin{tabular}{l| c| c}
			\hline
			& w/o stop-grad. & w/ stop-grad.\\
			\hline
			CIFAR-10 & 83.21$\pm$0.17  & \textbf{84.88$\pm$0.56}  \\
			
			Audio-MNIST & 97.45$\pm$0.29  & \textbf{99.41$\pm$0.21} \\
			\hline
		\end{tabular}
	\end{table}
	
	\subsubsection{the effect of the loss function}
	We compare the cosine distance with KL divergence and JS divergence in this part. For convenience, we utilize 'div' to represent divergence in the following table. Following Siamese \cite{chen2021exploring}, we apply stop-gradient to avoid collapsing solutions when minimizing the consistency loss (the effectiveness was analyzed in \ref{stop-grad}). 
	The results of table \ref{table4} show that the cosine distance is more suitable for the structure. 
	\begin{table}[H]\centering
		\renewcommand{\arraystretch}{1.4}
		\setlength{\tabcolsep}{1.4mm}
		\caption{The results of the effect of the loss function}\label{table4}
		\begin{tabular}{l| c| c |c}
			\hline
			& kl div & JS div & cosine distance\\
			\hline
			CIFAR-10 & 84.18$\pm$0.54 & 84.32$\pm$0.63 & \textbf{84.88$\pm$0.56} \\
			Audio-MNIST & 97.13$\pm$1.08 & 96.43$\pm$1.11 & \textbf{99.41$\pm$0.21}\\
			\hline
		\end{tabular}
		
	\end{table}
	\subsubsection{the effect of the consistency loss coefficient $\alpha$}
	To explore the effectiveness of the weight of $\alpha$, we increase the weight of consistent loss from 0.1 to 1.0. As shown in table \ref{table5}, As $\alpha$ increases, it works best on the Audio-MNIST dataset when it is equal to 0.2, and it has been decreasing since then. For the CIFAR-10 dataset, the effect is best when arises to 0.5, and then the accuracy gradually decreases. $ \alpha=0.2$ is more suitable for audio, $\alpha=0.5$ is more suitable for images. This means that $\alpha$ should be adjusted accordingly for different datasets. For different tasks, it is more appropriate to select different coefficients.
	\begin{table}[H]\centering
		\renewcommand{\arraystretch}{1.4}
		\setlength{\tabcolsep}{5mm}
		\caption{The effect of the consistency loss coefficient $\alpha$}\label{table5}
		\begin{tabular}{l|c| c}
			\hline
			$\alpha$& Audio-NMIST & CIFAR-10  \\
			\hline
			0.1 & 99.03$\pm$0.12 & 84.05$\pm$0.32 \\
			0.2 & \textbf{99.45$\pm$0.18} & 84.32$\pm$0.67 \\
			0.5 & 99.41$\pm$0.21 & \textbf{84.88$\pm$0.56} \\
			0.6 & 99.35$\pm$0.13 & 83.95$\pm$0.88 \\
			0.8 & 99.21$\pm$0.22 & 84.35$\pm$0.73 \\
			1.0 & 99.18$\pm$0.14 & 84.21$\pm$0.72 \\
			\hline
		\end{tabular}	
		
	\end{table}
	\section{conclusion}
	\vspace{0.3cm}
	We propose CR-Aug, a simple yet effective data augmentation regularization framework to address overfitting when training DL classifiers on image and audio datasets. Constraining the two augmented data by using cosine distance produces consistent predictions, and then using stop-gradient to mitigate collapsing solutions. Different data augmentation methods are given for different datasets. We verify the effectiveness and universality of the framework through experiments. This method is not limited to the above backbone and it can be easily applied to other models. So CR-Aug has huge potential in DL.
	
	In future work, we will further research the effectiveness of our algorithm in the medical dataset. It is known that medical images are harder to augment as compared to natural images due to the TB-consistent findings and the ROI spans for a relatively smaller portion of the whole image. We'll try to come up with a good CR-Aug data augmentation solution for medical images.
	\section*{Acknowledgement}
	This paper is supported by the Key Research and Development Program of Guangdong Province under grant No.
	2021B0101400003. Corresponding author is Shijing Si from Ping An Technology (Shenzhen) Co., Ltd (shijing.si@outlook.com).
	
	\footnotesize
	\bibliographystyle{IEEEtran}
	\balance
	\bibliography{refs}

\end{document}